\documentclass[journal]{IAENGtran}
\usepackage{wrapfig}
\usepackage{soul}
\usepackage{multirow}
\usepackage{booktabs}
\usepackage{adjustbox}
\usepackage{graphicx}
\usepackage{algorithm}
\usepackage[noend]{algorithmic}
\usepackage{xcolor,colortbl}
\usepackage{textcomp}
\usepackage{amsmath}
\usepackage[numbers,sort&compress,square]{natbib}
\definecolor{Gray}{gray}{0.85}
\definecolor{LightCyan}{rgb}{0.88,1,1}

\usepackage{epstopdf}
\usepackage{subfloat}
\usepackage{float}
\usepackage[font=small,labelfont=bf]{caption}
\usepackage[caption = false]{subfig}
\usepackage{caption}
\let\labelindent\relax
\usepackage{enumitem}
\setlist[itemize]{leftmargin=*}

\usepackage{lipsum} 
\usepackage{fancyhdr}
\begin{document}

\title{A New Classification Approach for Robotic Surgical Tasks Recognition}

\author{Mehrdad J. Bani, and~Shoele Jamali 
\thanks{Manuscript received June 26, 2017; revised June 26, 2017.}
\thanks{M. J. Bani and S. Jamali are with the Department of Computer Science, Shahid Beheshti University, Tehran, IRAN e-mail: (mehrdad4000@gmail.com).}
}

\maketitle

\begin{abstract}
Automatic recognition and classification of tasks in robotic surgery is an important stepping stone toward automated surgery and surgical training. Recently, technical breakthroughs in gathering data make data-driven model development possible. In this paper, we propose a framework for high-level robotic surgery task recognition using motion data. We present a novel classification technique that is used to classify three important surgical tasks through quantitative analyses of motion: knot tying, needle passing and suturing. The proposed technique integrates state-of-the-art data mining and time series analysis methods. The first step of this framework consists of developing a time series distance-based similarity measure using derivative dynamic time warping (DDTW). The distance-weighted $k$-nearest neighbor algorithm was then used to classify task instances. The framework was validated using an extensive dataset. Our results demonstrate the strength of the proposed framework in recognizing fundamental robotic surgery tasks.
\end{abstract}

\begin{IAENGkeywords}
Classification, Derivative dynamic time warping (DDTW), $k$-nearest neighbor, Robotic-assisted surgery, Task recognition.
\end{IAENGkeywords}

\thispagestyle{fancy}
\section{introduction}
\IAENGPARstart{T}he hospital operating room is a challenging work environment. Recently, some of these challenges have been addressed by introducing technological innovations such as Robotic Surgery \cite{lalys2014surgical, fard2017eee}, which promises to improve patient treatment by enabling shorter hospital stays, shortening recovery time and reducing the risk of infection. Current implementations operate in a tele-operation mode where the robotic surgery system relies exclusively on direct surgeon input. 

Future advances will automate more aspects of robotic surgery procedures \cite{pandya2014review, fard2016toward}. It is, however, quite clear that to develop such an autonomous systems, a more rigorous model of surgical procedures is needed. Surgical motions need to be modeled and quantified to make them amenable for further study. Goal-oriented human motion and human language are analogous as both of them consist of a low-level elements that, when combined in meaningful sequences, result in an emergent meaning or higher-level task. Hence, techniques that have effectively been applied in the analysis of human speech and language are natural candidates to apply to surgical motion modeling. Consequently, the ``{\em Language of Surgery}'' has been defined as a systematic description of surgical activities and rules for decomposition \cite{gaojhu}. More specifically, the language of surgical motion includes describing particular activities that are performed by surgeons with their instruments or hands to accomplish a planned surgical objective. 
Current systems like {\em da Vinci} (Intuitive Surgical, Sunnyvale, CA) \cite{guthart2000intuitivetm} record motion and video data, enabling development of computational models to recognize and analyze surgical performance through data-driven approaches. Recent advances in data mining research for uncovering concealed patterns in huge dataset, like kinematic and video data, offer the possibility to better understand surgical procedures from a system point of view. Thus, the key step for advance research in surgical task recognition is to develop techniques that are capable of accurately recognizing fundamental surgical tasks such as suturing, knot tying and needle passing.

In this paper we extend the \cite{RCS:RCS1766} by present a new framework to classify robotic-assisted surgical tasks based on Derivative Dynamic Time Warping (DDTW) with the well-known distance-weighted $k$-nearest neighbor ($k$NN) classification method.

\vspace{1em}\section{related work}
In recent years, recognizing and understanding surgical procedures at different levels of granularity has been a focus of research \cite{jahanbani2016computational, reiley2011review}. Surgical procedures can be generally broken down to four main levels, from higher to lower: phases, steps, tasks and motions \cite{lalys2014surgical}. At the higher level of surgical process modeling, statistical models have been proposed using recorded force and motion data \cite{rosen2001markov, rosen2002task}, surgical tool usage \cite{Blum2008} and video data \cite{Lalys2012} to classify surgery phases. Most existing work has addressed the recognition of activities using different techniques such as neural networks and Hidden Markov Models (HMM)  \cite{sanchez2008activity, doi:10.1108/IJQRM-05-2011-0075}. At the lower level, effort has been applied to detect surgical motion  \cite{lin2006towards, Reiley2008} or model surgical gestures and classify them using different methods such as HMM and Linear Discriminant Analysis (LDA) \cite{reiley2009task}. A common drawback in these methods is that they are time consuming and require significant human interaction and pre-processing. 

While many of the studies in the literature focused on detecting surgical motion at the more granular level \cite{padoy2012statistical, Zappella2013}, developing quantitative classification techniques that can be used as a framework to differentiate important tasks during surgical procedures needs to be investigated. Here, a task is defined as a sequence of activities used to achieve a surgical objective \cite{lalys2014surgical}. This work focuses on three fundamental tasks during robotic-assisted minimally invasive surgery: suturing, knot tying and needle passing. These tasks are commonly part of a surgical skills training curriculum \cite{gaojhu}. With the advent of robotic surgery devices, a huge amount of data, including temporal kinematic signal, can be captured during surgeries. Our work seeks to take this information and build a framework to recognized three main robotic surgery tasks by measuring similarities between their temporal data and underlying signatures. 

Dynamic Time Warping (DTW) \cite{bernad1996finding} is a well-known technique for time series classification \cite{fu2011review}. In the surgical procedure application, it has been used to classify surgical processes \cite{forestier2012classification} and surgical gestures \cite{Lalys2012}. While DTW has been successfully used in many domains, it may however, fail to find the obvious natural alignments between two sequences when they have significant difference in their signal function over time. Thus, Derivative Dynamic Time Warping (DDTW) was proposed in \cite{Keogh2001} and it has been shown to provide promising results to address this issue. The similarity that has been derived from DTW or DDTW, can be used as an input to the $k$-Nearest Neighbors algorithm ($k$NN), a popular classification method, to classify a new data based on its similarity to other sample data \cite{7564399,bhatia2010survey}. 

The main focus of our work is to investigate the feasibility of task classification during robotic-assisted surgery \cite{fard2016machine}. This is in contrast to most work in this domain, which used video data or observation-based methods \cite{Lalys2012}. We develop distance-weighted $k$NN classification method using similarity measure derived from DTW and DDTW. Our work differs from previous studies in the sense that we use only Cartesian data of both right and left hand tool position with minimum pre-processing that results in simple, straightforward and accurate framework. 
\vspace{-1.2em}

\renewcommand{\vec}[1]{\mathbf{#1}}
\section{method}
\begin{figure}[t]
\centering
\scalebox{0.33}{\includegraphics{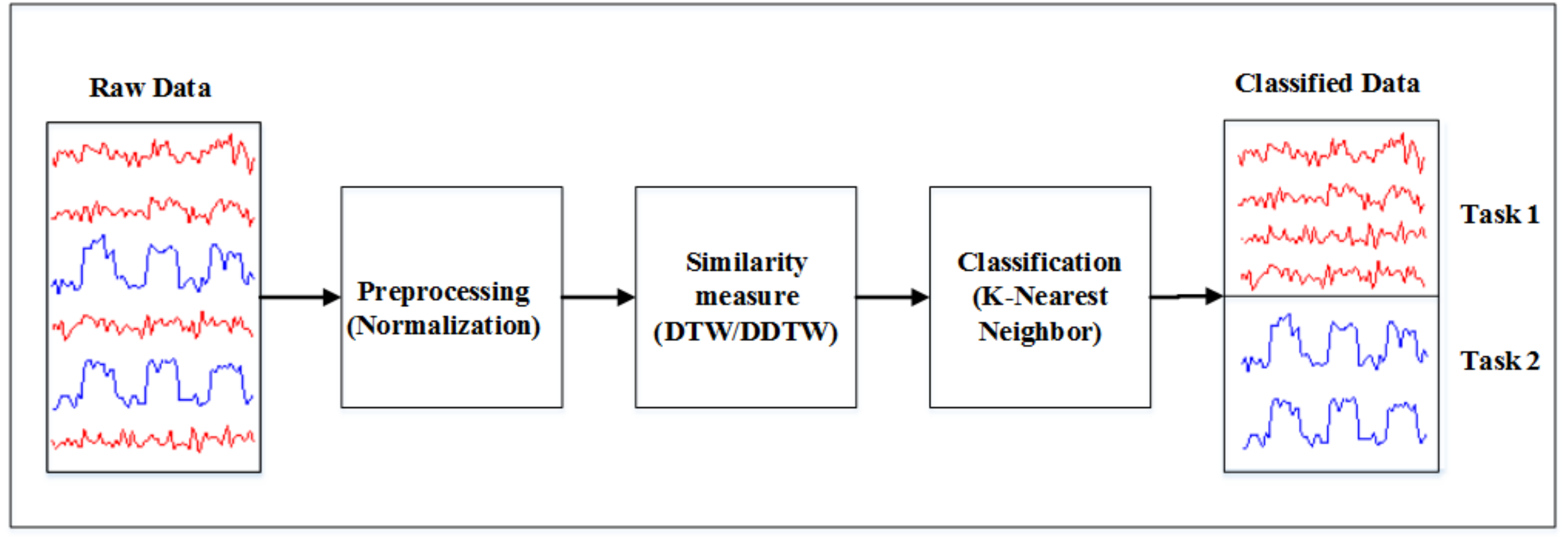}}
\caption{Proposed framework consists of three steps: preprocessing, similarity measurement between time series data, and classification using $k$-nearest neighbor method.}
\label{flowchart}
\end{figure}

The aim of our work is to recognize robotic surgery tasks. As noted before, this work focus on three important fundamental robotic surgery tasks: knot tying, needle passing and suturing. These tasks are part of a fundamental laparoscopic surgery (FLS) skills training program \cite{fried2004proving, fard2016early}. The classification framework that is developed in this study, contains of three key components. The first component is quantitative measures of the different tasks. We analyze motion data from robotic surgery device to extract multivariate time series datasets that represent different tasks. After preprocessing and normalization of data, the subsequent step is measuring the similarity between different surgical tasks. In this study we employ DDTW to measure similarity between multidimensional time series data. The third component is the classification algorithm, which is based on the distance-weighted $k$-nearest neighbor approach. The combination of these three steps results in a novel task classification framework for robotic surgery data. Figure \ref{flowchart} shows the summary of our proposed framework. In the following sections, each step in the framework will be discussed in detail. 
\vspace{-1.5em}

\subsection{Quantification of robotic surgery Tasks}

In this study, we implement our model using ``JHU-ISI Gesture and Skill Assessment Working Set (JIGSAWS)'' \cite{gaojhu} where data have been gathered using a {\em da Vinci} surgical system \cite{guthart2000intuitivetm}. This surgical activity data includes different sorts of robotic-assisted surgery features, such as surgeon kinematic and video data during surgery procedures that has been captured by an Application Programming Interface (API). Using a {\em da Vinci}, a surgeon operates passive master tool manipulators (MTMs), directing resultant teleoperated movement in active patient-side manipulators (PSMs). Time series data for each of the robot arms (MTMs and PSMs) has been gathered for three fundamental tasks: knot tying, needle passing and suturing.
\vspace{-1em}

\subsection{Similarity Measures}
\begin{figure}[h]
\centering
\subfloat[]{\scalebox{.23}{\includegraphics{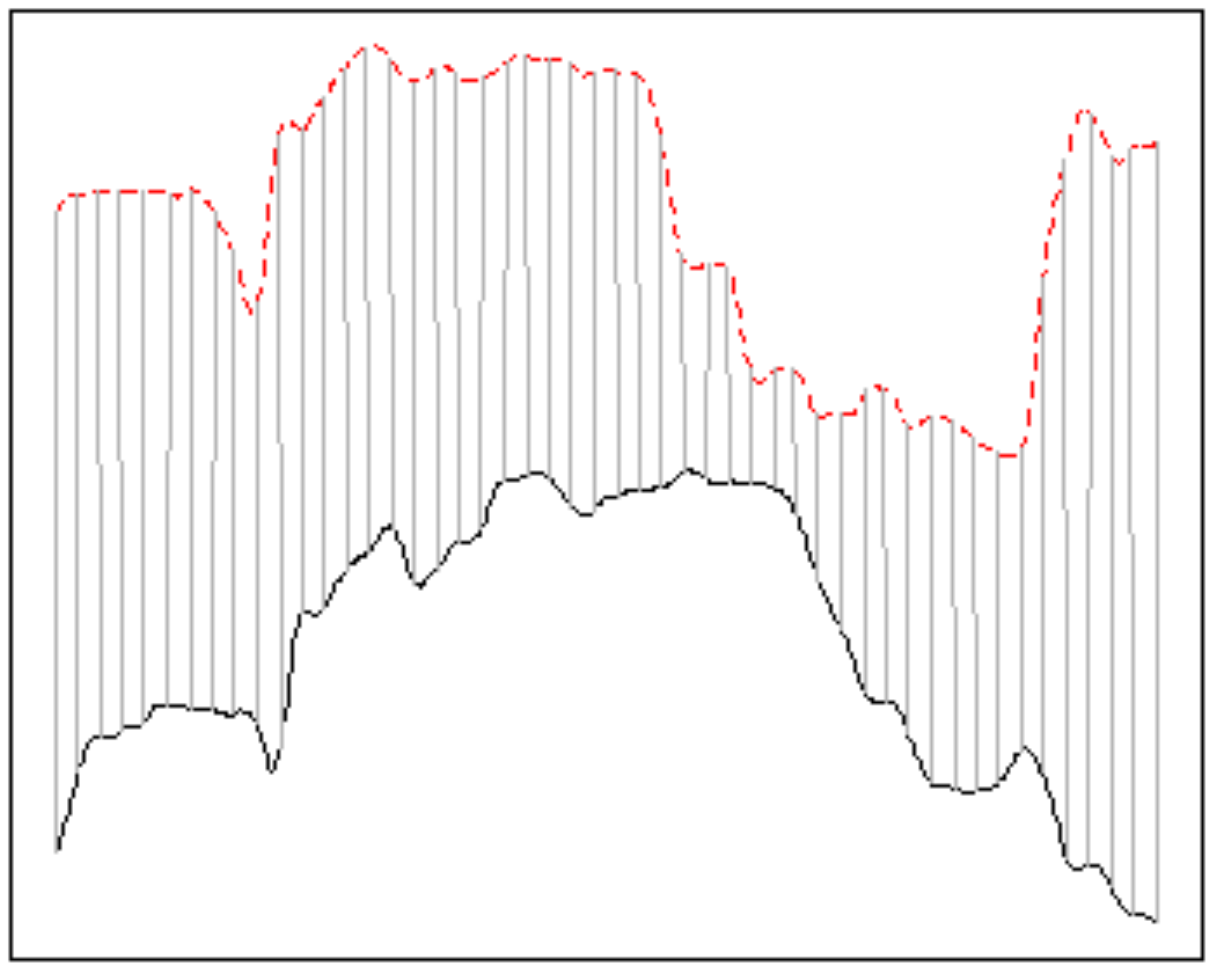}}}
\subfloat[]{\scalebox{.23}{\includegraphics{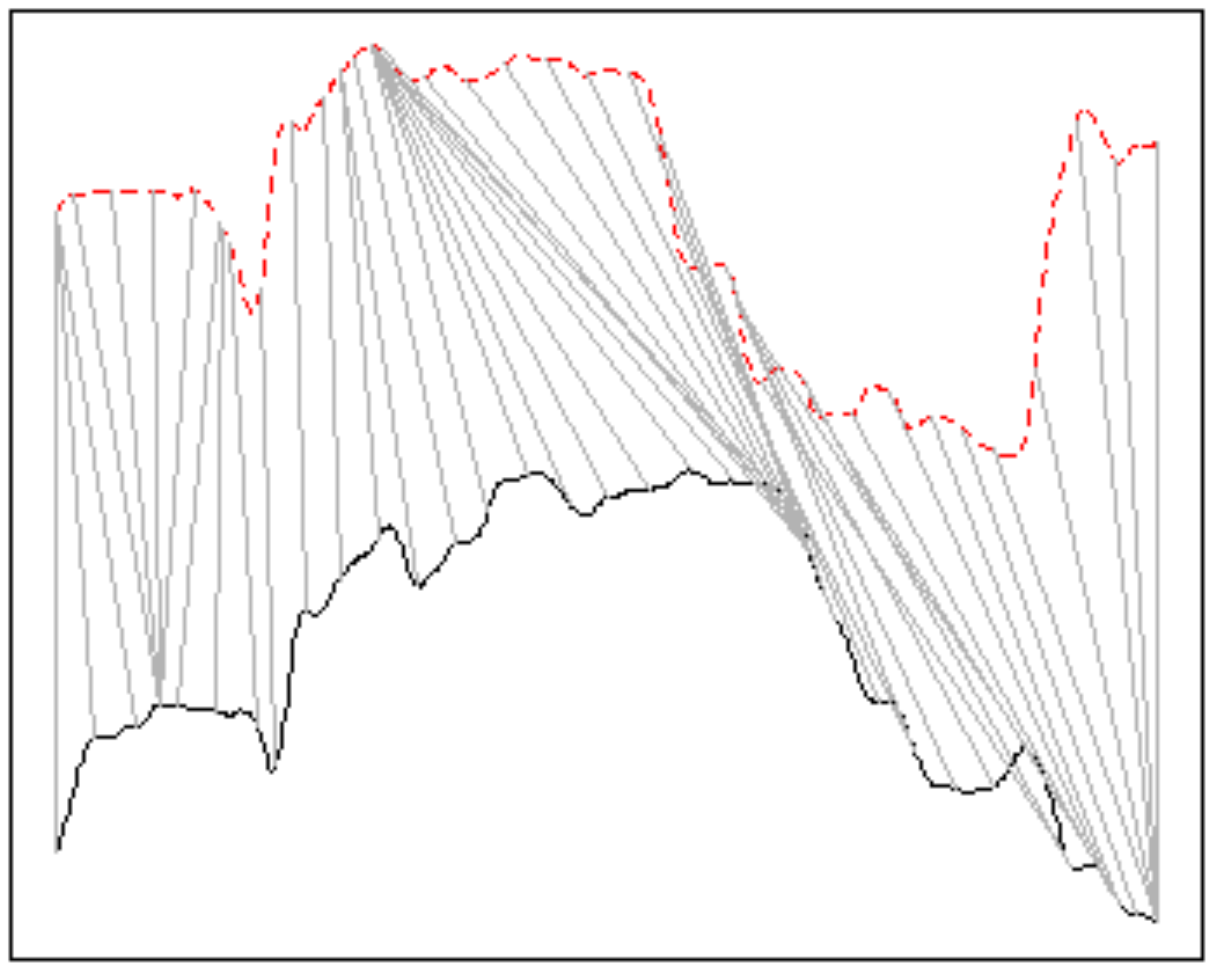}}}
\subfloat[]{\scalebox{.23}{\includegraphics{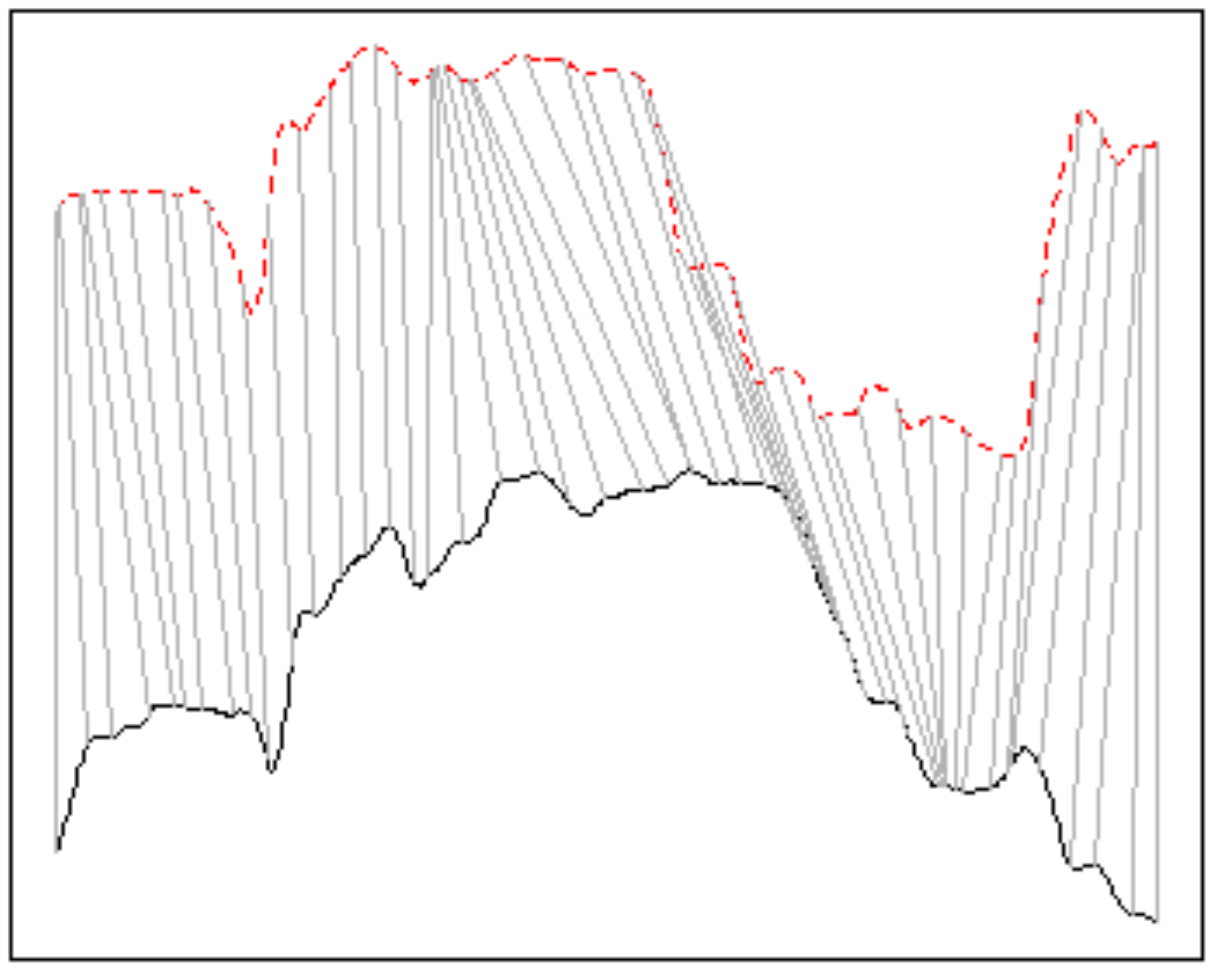}}}
\caption{Comparison between a) Euclidian distance b) DTW and c) DDTW for X-axis time series of Knot Tying trial 11 and 12 in JIGSAWS dataset.} 
\label{comp}
\end{figure}

The choice of method for measuring (dis)similarity is a critical step in achieving valid classification results. One of the primary issues to measure the similarity between two time series using a distance measurement methods such as Euclidean distance is that the outcomes can, in some cases, be exceptionally unintuitive due to sensitivity to distortion in the time axis (Fig. \ref{comp}). If, for instance, two time series are indistinguishable, however slightly out of phase with one another, then a distance measure such as the Euclidean distance will give an extremely poor similarity measure. Dynamic Time Warping (DTW) has been developed to overcome this problem \cite{bernad1996finding}. In this work, we propose a novel implementation of {\em DTW} and a related method, {\em DDTW}, for time series data of robotic surgery tasks. 
\vspace{0.4em}

\begin{figure}
\centering
\scalebox{0.5}{\includegraphics{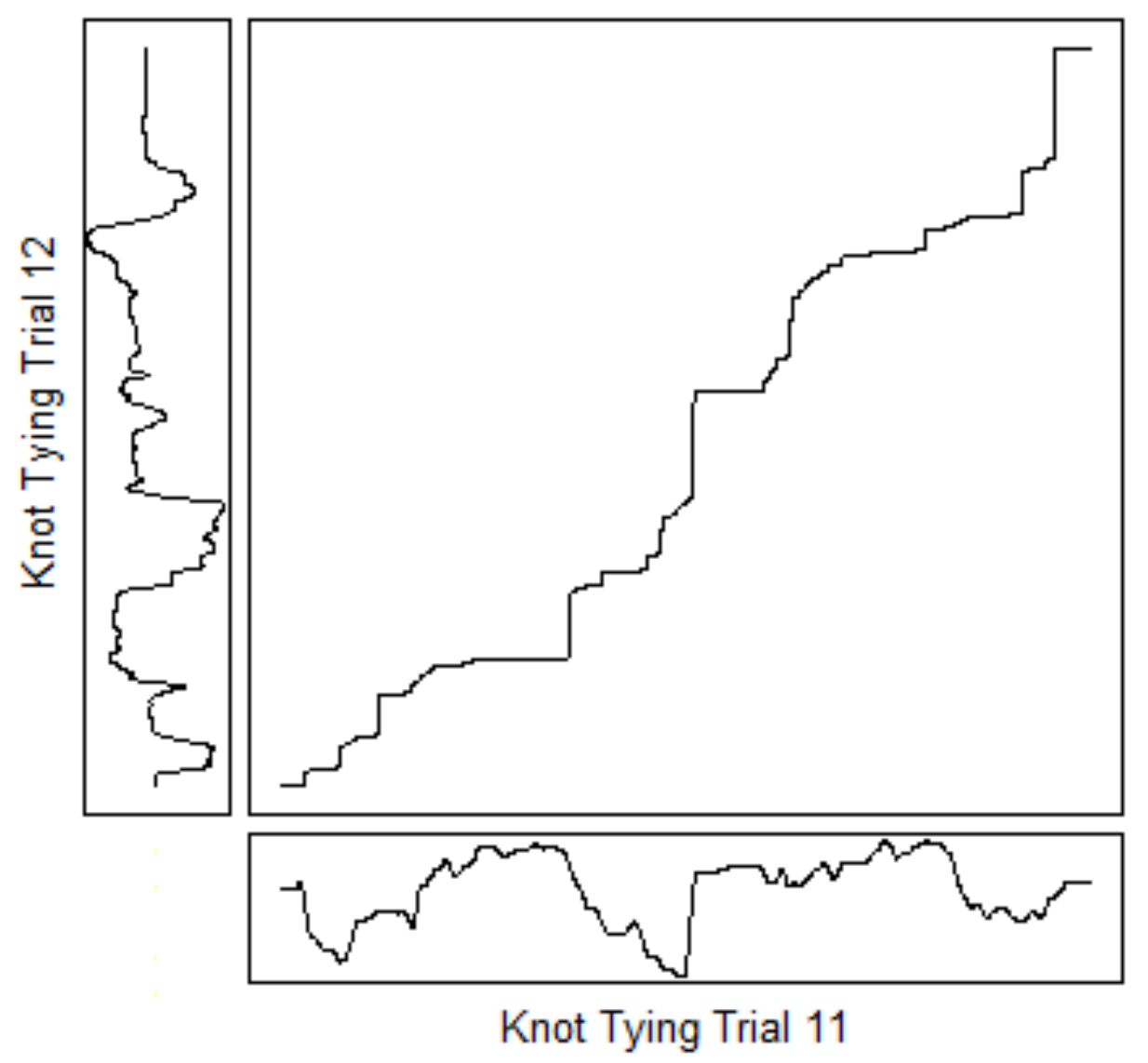}}
\caption{Time series alignment using warping matrix with the minimum distance warp path of X-axis time series of two Knot Tying trials (11 and 12) in JIGSAWS dataset.}
\label{warp-matrix}
\end{figure}

Dynamic time warping is a common approach to measure the dissimilarity between two sets of time series data, even if the lengths of the time series do not match. DTW can find an optimal alignment between two time-dependent sequences under specific constraints. Essentially, the sequences are warped in a nonlinear fashion to match each other. Given two $p$-dimensional time series $\vec{S} =(\vec{s_1}, \vec{s_2}, ..., \vec{s_m})$ and $\vec{T} =(\vec{t_1}, \vec{t_2}, ..., \vec{t_n})$ where $\vec{S}$ and $\vec{T}$ have $m \times p$ and $n \times p$ dimension respectively, these two sequences can be arranged as $m \times n$ matrix like the sides of a grid (Fig. \ref{warp-matrix}) in which the distance between every possible combination of time instances $\vec{s_i}$ and $\vec{t_j}$ is stored. Both sequences start on the bottom left of the grid. 
For multidimensional DTW, we use the well-known Euclidean distance measure to find a distance between two $p$-dimensional sequences (Eq. \ref{ED}). 
\vspace{-1.2em}

\begin{equation}
d(\vec{s_i},\vec{t_j})=\sqrt{\sum_{l=1}^{p}{(s(i,l)-t(j,l))^2}}
\label{ED}
\end{equation}

To find the best match between two sequences, a path through the grid that minimizes the overall distance between them is needed. In order to compute overall distance, all possible routes through the grid must be found. Then, the overall distance is calculated to be the minimum of the sum of the distances between the individual elements on the path divided by the sum of the weighting function. It is evident that for long sequences the quantity of conceivable ways through the network will be very large. Several constraints such as monotonicity, continuity, boundary, slope constraint and warping window constraint apply to limit the moves that can be produced from any point in the path. Among those, warping window which can be defined as subset of the matrix that is available, should be provided as an input parameter to the model. 

The power of the DTW algorithm is that rather than exploring every conceivable path through the grid keeps track of the cost of the best path. Thus, DTW distance can be formulated as a dynamic programming problem. Using a dynamic programming approach, the warp path must either be incremented by one unit or stay at the same $i$-axis or $j$-axis. Therefore, one can formulate it as recurrence of cumulative distance, defined as: 
\vspace{-0.7em}
\begin{equation}
DTW(i,j) = d(\vec{s_i},\vec{t_j})+DTW_{min}
\label{dtw}
\end{equation}

where $d(\vec{s_i},\vec{t_j})$ can be calculated using Equation (\ref{ED}) and $DTW_{min}=min\{DTW (i,j-1), DTW(i-1,j), DTW(i-1,j-1)\}$.
\vspace{0.4em}
DDTW is a modification of DTW to consider higher-level features of a sequence\textquotesingle s shape instead of Y-axis values of data points. 
In some application when a feature such as peak or valley in one sequence is little higher or lower than corresponding feature in another sequence, DTW may neglect to discover this type of alignment (Fig. \ref{comp}) 
\cite{fu2011review}. Thus DTW may fail to find obvious natural alignments between time series data of two instances of the same sequence. 
To address this issue, Derivative Dynamic Time Warping (DDTW) algorithm has been proposed \cite{Keogh2001}. 
The proposed framework adapts DDTW by taking the first derivative of the sequence of time series data for different robotic surgery tasks. Considering simplicity and generality, the following estimate for the derivative of each point in time series is used: 

\begin{equation}
D_i[s]=\frac{(s_i-s_{i-1})+(\frac{s_{i+1}-s_{i-1}}{2})}{2}
\label{ddtw}
\end{equation}

This estimate is the average of the slope of the line through the point $\vec{s_i}$ and its left neighbor $\vec{s_{i-1}}$, and the slope of the line through the left neighbor $\vec{s_{i-1}}$ and the right neighbor $\vec{s_{i+1}}$. 
Like DTW, an $m \times n$ matrix is constructed that contains the distance between $\vec{s_i}$ and $\vec{t_i}$ using the square of difference of $D_i[s]$ and $D_j[t]$, the estimated derivative of $\vec{s_i}$ and $\vec{t_j}$. 
\vspace{-1em}

\subsection{Weighted $k$-Nearest Neighbor Classification}
The $k$-Nearest Neighbors algorithm ($k$NN) is a non-parametric instance-based method used for classifying a new data based on the majority label of its $k$ nearest neighbors in the training set \cite{bhatia2010survey}. 
The most significant difference between instance-based classifiers and other classification methods is that unlike other sophisticated methods in this domain, it does not require $a$ $priori$ knowledge of underlying patterns in data.
It is intuitive that observations which are close together based on some appropriate metric will have the same class label. Thus, simplicity, effectiveness, intuitiveness and accuracy of $k$NN suggests its use in many areas. 
A refinement of this classification algorithm is distance-weighted $k$NN in which each of the $k$ neighbors weight the evidence of a neighbor close to an unclassified observation more heavily than others with the greater distance to the query observation \cite{dudani1976distance, mahtabIEOM2015}. Let us define the $k$ nearest neighbor of query $x_q$ as $D_{kNN}=\{(x_i,y_i); i=1,...,k\}$ and $d_i$ as a distance between $i$th nearest neighbor and $x_q$. Then a weight $w_i$ attributed to $i$th nearest neighbor can be defined as

\begin{equation}
w_i= \frac{d_k-d_i}{d_k-d_1}, d_k \neq d_1
\label{wknn}
\end{equation}

Thus, the classification result of the query can be made as
\begin{equation}
y_q=\underset{y}{\arg\max}\sum_{(x_i,y_i)\in D_{kNN}}{w_i\times \delta(y=y_i)}
\label{wknn-class}
\end{equation}

According to the Eq. (\ref{wknn}), a neighbor with smaller distance has more weight than the one with greater distance.
The balance of simplicity on one hand and accuracy on the other hand led us to choose this method for our time series robotic surgery task classification. The only parameter that needs to be provided is $k$. In general, a small value of $k$ means that any noise present with the data will have a higher influence on the result, however, a large value lets the samples of the other classes get included in the neighborhood of test data, resulting in poor classification and high computational expense. In order to find the best value of $k$ to maximize the classification performance, we will train the model by examining the accuracy as a function of $k$.
\vspace{-1.5em}

\section{experimental setup}
In this section, we will describe the dataset that is used for evaluating the proposed time series classification framework for robotic-assisted surgical tasks along with detail of implementation and performance evaluation. \vspace{-1.5em} 

\subsection{Dataset Description}
\begin{figure}[h]
\centering
\subfloat[ ]{\scalebox{.35}{\includegraphics{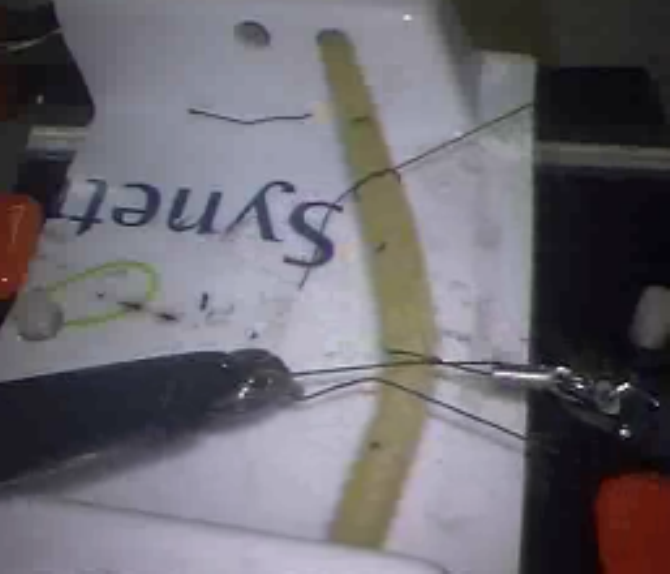}}}
\subfloat[ ]{\scalebox{.35}{\includegraphics{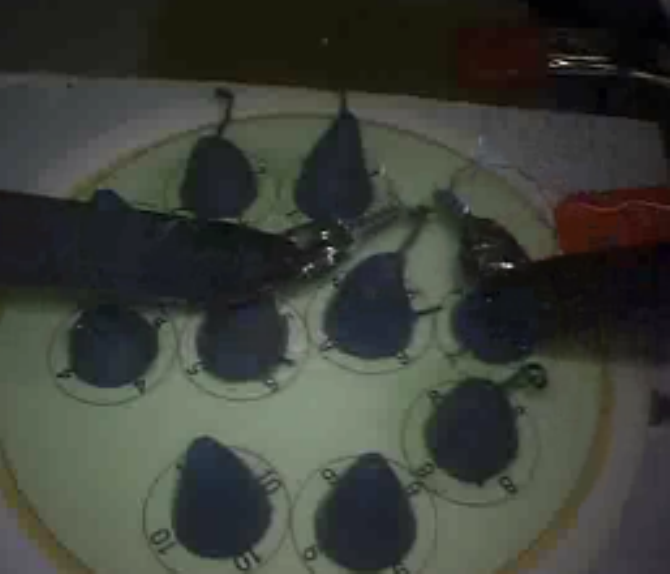}}}
\subfloat[ ]{\scalebox{.35}{\includegraphics{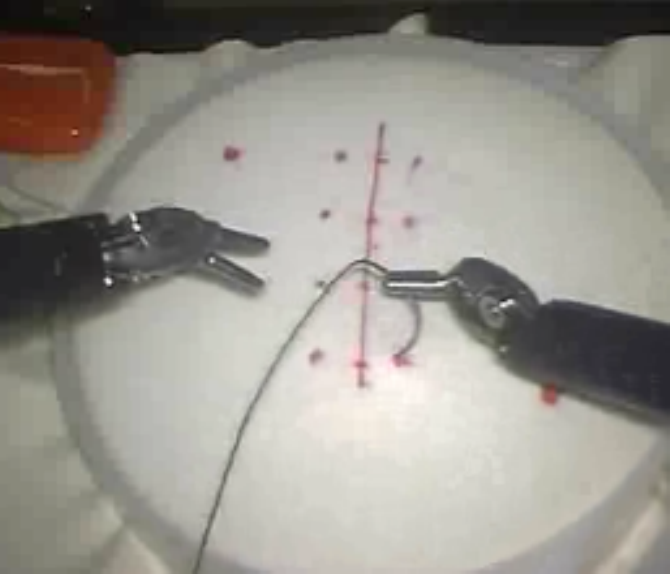}}}
\caption{Three fundamental robotic surgery task: a) Knot Tying, b) Needle Passing and c) Suturing}
\label{fig:3task}
\end{figure}

As briefly explained before, we are using the 
JIGSAWS dataset \cite{gaojhu}. JIGSAWS is comprised of data for three fundamental surgical tasks performed by surgeons (Figure \ref{fig:3task}). These tasks include: 

For each of the three tasks, we analyze kinematic data captured using the API of the {\em da Vinci} at 30 Hz. The data includes 19 kinematic variables for Cartesian position, rotation matrix, linear velocities, angular velocities and a gripper angle. The left and right MTMs, and the left and right PSMs are included in the 76-dimensional dataset.
We build our model using 3D Cartesian position ($x,y,z$) data from both right and left PSMs. The JIGSAWS includes data from eight right-handed surgeons where all of them repeated each surgical task five times (i.e. trials) \cite{gaojhu}.
\vspace{-1.1em}

\subsection{Implementation Details}
Separating data into training and testing sets is a vital step of any model evaluation. For classification methods, the training set is used to discover initial patterns in data, while the testing set helps us evaluate whether or not the recognized patterns hold. 
One of the popular methods in this regard is stratified $n$-fold cross validation with an equal proportion of classes in each fold to reduce the bias of training and test data \cite{kohavi1995study}. 
In each run, $n$-1 out of $n$ folds are used for training and the remaining one fold is used for testing. We chose the widely accepted 10-fold cross validation method, and for the sake of comparison also used the Leave-One-Out (LOO), which is a special case of $n$-fold cross validation when $n$=$N$ and $N$ is the total number of data points. In each fold all but one observation is used for training and the left out observation is tested. 
One hundred replications were conducted for each method to get more robust results 
and the average and standard deviation is reported in the results section. 
It should also note that in preliminary analysis the choice of different warping window size does not affect the results significantly. Hence, we set the window size to 100 for all analyses which resulted in minimum parameter tuning for the DTW method \cite{ratanamahatana2005three}.
\vspace{-1.2em}

\subsection{Performance Evaluation}
The first step in performance evaluation is to tabulate the results of all classifications into a 
corresponding confusion matrix (Table \ref{tab:conf-matrix}). The correctness of a classification can be assessed by calculating the number of correctly classified instances, called true positives (TPs). True negatives (TNs) are the number of correctly classified instances that do not belong to the class. If a data point is incorrectly assigned to the class it is a false positive (FPs), and if it is not classified as class instances it is a false negative (FNs). 
\vspace{.5em}
\begin{table}[htbp]
\centering
\caption{Illustration of True Positive, False Positive, True Negative, and False Negative for class Task 1, using a multi-class confusion matrix.}
\begin{tabular}{|c|c|c|c|c|}
\cline{3-5} 
\multicolumn{1}{c}{} & & \multicolumn{3}{c|}{\textbf{Predicted}}\tabularnewline
\cline{3-5} 
\multicolumn{1}{c}{} & & Task 1 & Task 2 & Task 3\tabularnewline 
\hline 
\multirow{3}{*}{\textbf{Actual}} & Task 1 & {\scriptsize{}TP} & {\scriptsize{}FN} & {\scriptsize{}FN}\tabularnewline
\cline{2-5} 
& Task 2 & {\scriptsize{}FP} & {\scriptsize{}TN} & {\scriptsize{}FN}\tabularnewline
\cline{2-5} 
& Task 3 & {\scriptsize{}FP} & {\scriptsize{}FN} & {\scriptsize{}TN}\tabularnewline
\hline 
\end{tabular}
\label{tab:conf-matrix}%
\end{table}%
\vspace{0.5em}

Based on the values in the confusion matrix, different classification performance measurements are widely used, such as accuracy, sensitivity and specificity. Accuracy measures the fraction of correctly classified data. Sensitivity measures proportion of positive instances that are classified as positive and specificity measures proportion of negative instances that are classified as negative. In this paper we evaluate our classification framework for three class of tasks \cite{sokolova2009}. 
One important factor to consider is the number of data points in each class. Since we do not have an equal number of cases of each task, we modified the measurement for multi-class classification by adding $\rho_i$ which is $\frac{n_i}{N}$ and $n_i$ is number of instances in class $i$ and $N$ is total number of instances in dataset.
\vspace{-0.7em}
\begin{equation}
Accuracy=\sum_{i=1}^{C}{\rho_i\frac{TP_i+TN_i}{TP_i+FN_i+FP_i+TN_i}}
\label{accu}
\end{equation}
\vspace{-.7em}
\begin{equation}
Sensitivity=\sum_{i=1}^{C}{\rho_i \frac{TP_i}{TP_i+FN_i}}
\label{sen}
\end{equation}
\vspace{-.7em}
\begin{equation}
Specificity=\sum_{i=1}^{C}{\rho_i \frac{TN_i}{FP_i+TN_i}}
\label{spe}
\end{equation}
where in above equations, C refers to number of classes.
\vspace{-.3em}

\section{Results and Discussion}
In this section we provide the experimental results from using the proposed classification framework on three robotic surgery tasks. Two similarity methods, DTW and DDTW, were used to measure the pairwise distance between tool-tip paths during surgical tasks. Before that we need to train our classifier for the best value of $k$-nearest neighbors.

\begin{figure}[t]
\centering
\scalebox{.6}{\includegraphics{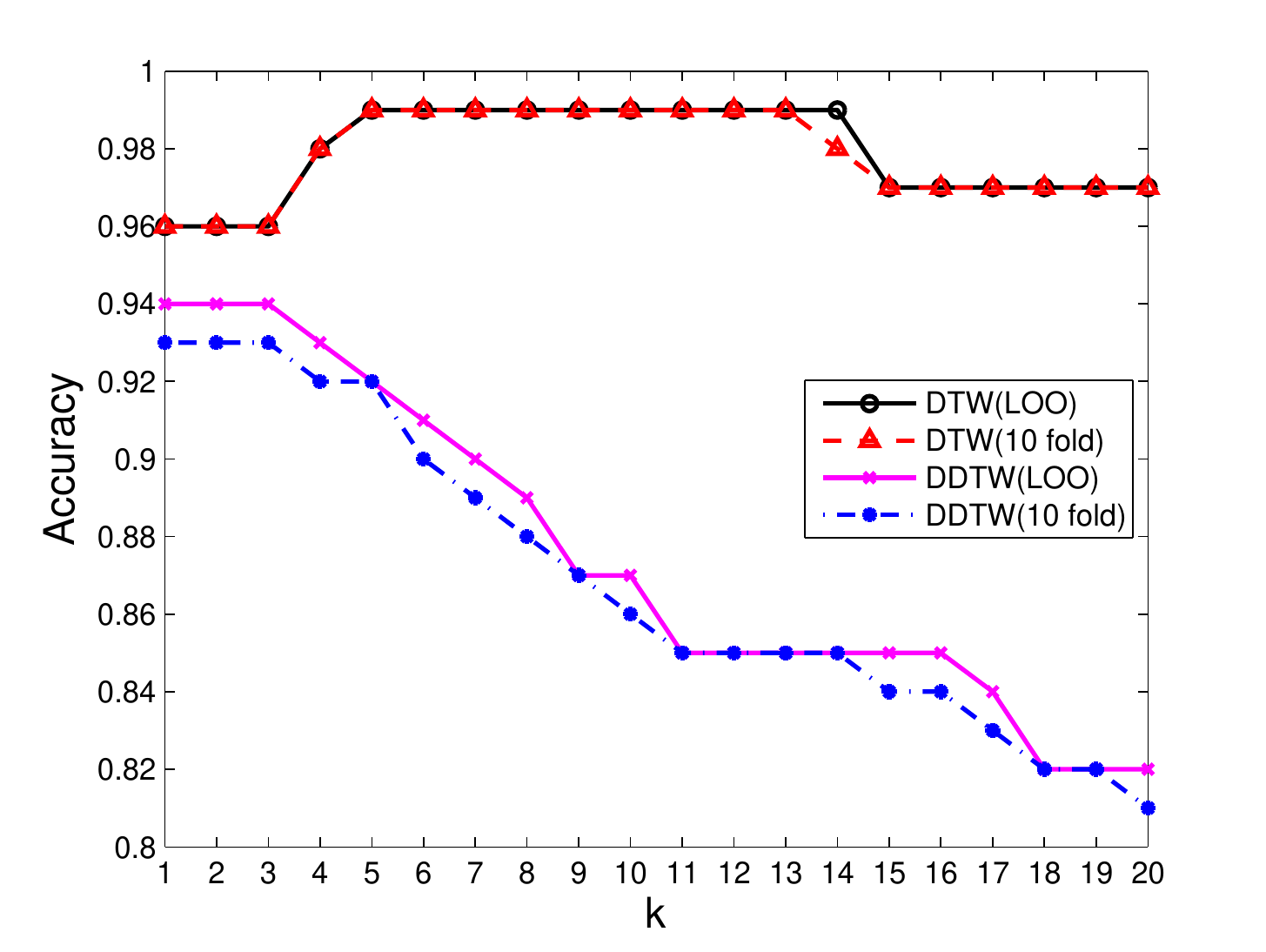}}
\caption{Comparison of accuracy for different similarity measures and validation techniques as a function of $k$}
\label{fig:knn}
\end{figure}

Figure \ref{fig:knn} represents the accuracy as a function of $k$ 
using different similarity measures and validation techniques. As mentioned earlier, 10-fold and LOO cross validations are employed in this study. We can observe that the $k$NN classification using DTW similarity measurement achieves its best performance when $k$=6 for both 10-fold and LOO cross validation techniques. However, DDTW best performance achieves when $k$=3. 
Also it can be observed that DDTW is more sensitive to the value of $k$ compared to DTW. This may be due to the smoothing properties of the derivative, which may mask unique features in the data that would be required to distinguish a task among a larger number of potential classes.
\vspace{0.5em}
We obtain the result of accuracy as function of $k$ to identify the best classification scheme on robotic surgery task dataset. The accuracy of the best scheme for three different scenarios of using only right hand, left hand or both data with 10-fold and LOO cross validations are listed in Table \ref{tab: all}. 
For two-handed Cartesian data, DTW-$k$NN achieved a top accuracy of 99.4\%, while for DDTW-$k$NN the highest accuracy was 93.6\%.

From Table \ref{tab: all}, DTW is shown to consistently out-perform DDTW. This implies that the DTW method is capable of capturing specific patterns in surgical tool tip time series path. Thus, despite the promising result from DDTW in other domains, it might not give a higher accuracy compared to DTW for robotic surgery data. This will lead us to conclude that the local differences in position of surgical device tool tip over time for each task is very important. Knowing that DDTW is designed to not be sensitive to sudden peaks and valleys (compared to DTW), we can conclude that these peaks and valleys are a meaningful feature of robotic surgery tasks. Removing those features resulted in losing some information required for proper classification. 
\begin{table}[t]
\centering
\caption{Comparison between accuracy using DTW and DDTW method and different validation techniques for right, left and both hand movement path data (and their standard deviation).}
\vspace{0.3em}
\begin{adjustbox}{width=0.47\textwidth}
\begin{tabular}{|c|c|c|c|c|c|c|}
\cline{2-7} 
\multicolumn{1}{c|}{} & \multicolumn{2}{c|}{\textbf{Right Hand}} & \multicolumn{2}{c|}{\textbf{Left Hand}} & \multicolumn{2}{c|}{\textbf{Both Hands}}\tabularnewline
\cline{2-7} 
\multicolumn{1}{c|}{} & DTW & DDTW & DTW & DDTW & DTW & DDTW\tabularnewline
\hline 
\multirow{2}{*}{\textbf{10-fold}} & 97.76\% & 93.20\% & 98.58\% & 89.10\% & 99.28\% & 93.06\%\tabularnewline
& (0.56\%) & (0.89\%) & (0.57\%) & (0.85\%) & (0.29\%) & (0.90\%)\tabularnewline
\hline 
\multirow{2}{*}{\textbf{LOO}} & \textbf{98.09\%} & 93.98\% & \textbf{98.80\%} & 89.30\% & \textbf{99.40\%} & 93.60\%\tabularnewline
& \textbf{(0.00\%)} & (0.15\%) & \textbf{(0.00\%)} & (0.17\%) & \textbf{(0.00\%)} & (0.09\%)\tabularnewline
\hline 
\end{tabular}
\end{adjustbox}
\setlength{\tabcolsep}{14pt}
\label{tab: all}%
\end{table}%
It is worth noting that the $k$NN classification method is sensitive to training set size. The size of the training set increases with a higher number of folds in $n$-fold cross validation. Consequently, we would expect $k$NN to perform better with LOO compared to 10-fold in terms of both better accuracy and lower standard deviation (Table \ref{tab: all}).

\begin{figure}[h]
\centering
\scalebox{.32}{\includegraphics{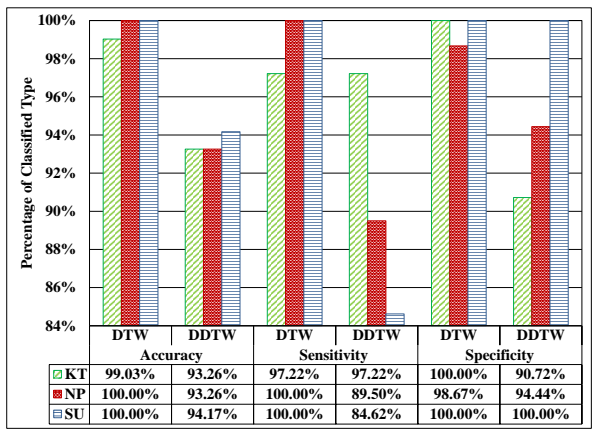}}
\vspace{-2.2em}
\caption{Comparison of performance measure using DTW and DDTW method with LOO cross validation and best $k$-NN classification for different task: Knot Tying (KT), Needle Passing (NP) and Suturing (SU)}
\label{fig:all}
\end{figure}

Figure \ref{fig:all} compares DTW and DDTW for each surgical task using data from both tool tips. It clearly shows that DTW gives the best performance for all tasks. All suturing and needle passing tasks can be correctly classified while only one of the knot tying is misclassified as needle passing. Also, knot tying and suturing have the best specificity, which means that fewer tasks were misclassified as suturing or knot tying. One can conclude that all these tasks have the unique features that make them recognizable among different surgeons with different expertise. 
\vspace{-1.3em}
\vspace{-1em}

\section{conclusion}
In this study we pursued the open question of the classifiability of fundamental surgical tasks in robotic-assisted minimally invasive surgery. We proposed a three-step classification framework for RMIS task recognition. Our method analyzes motion trajectory data obtained from the API of a {\em da Vinci} robotic surgery device. We developed distance-weighted $k$-nearest neighbor classification approach that use similarity measures obtained from DTW and DDTW for each task. The performance of the proposed framework based on the experimental results are encouraging with 99.4\% accuracy. This result establishes the feasibility of applying time series classification methods on RMIS tool tip position data to recognize the three fundamental tasks during robotic minimally invasive surgery (i.e., suturing, knot tying and needle passing). A key advantage of our approach is its simplicity by using only 3D Cartesian movement path of the right and left hand tool tips. Despite the high accuracy that achieved in this study, DTW has polynomial time complexity $O(N\times n^2)$ where $N$ is the number of sample in the data and $n$ is the length of time series. Thus, the proposed method might not be very efficient as an option when quick task classification is desired.
Therefore, future work should investigate for more computationally efficient methods to measure similarity between motion paths.

Furthermore, reliable classification is possible in light of the fact that time series features of these three tasks are differentiable from each other. This approach can be applied in a straightforward manner for development of an online gesture recognition system during robotic-assisted surgery. It can also facilitate robotic surgical skill assessment and training curriculum \cite{RCS:RCS1850}. Perhaps most excitingly, this framework can lay the groundwork towards development of semi-autonomous robot behaviors, such as automatic camera control during robotic-assisted surgery by detecting the task that is being performed. However, a prior step to that is to test the performance of our model in a real surgical environment in the present of other tasks or possible noise. Thus, there may be utility in extending our work by adding noise or other tasks (beside those in the training set) to the data in order to build a more robust task recognition method.

\bibliographystyle{BibTeXtran}   
\small \bibliography{refIEEE}       

\end{document}